\documentclass[letterpaper]{article} % DO NOT CHANGE THIS
\usepackage{aaai2026}  % DO NOT CHANGE THIS
\usepackage{times}  % DO NOT CHANGE THIS
\usepackage{helvet}  % DO NOT CHANGE THIS
\usepackage{courier}  % DO NOT CHANGE THIS
\usepackage[hyphens]{url}  % DO NOT CHANGE THIS
\usepackage{graphicx} % DO NOT CHANGE THIS
\usepackage{natbib}  % DO NOT CHANGE THIS AND DO NOT ADD ANY OPTIONS TO IT
\usepackage{caption} % DO NOT CHANGE THIS AND DO NOT ADD ANY OPTIONS TO IT
\usepackage{algorithm}
\usepackage{algorithmic}
\nocopyright
\usepackage{newfloat}
\usepackage{listings}
\DeclareCaptionStyle{ruled}{labelfont=normalfont,labelsep=colon,strut=off} % DO NOT CHANGE THIS
\floatstyle{ruled}
\newfloat{listing}{tb}{lst}{}
\floatname{listing}{Listing}
\usepackage[absolute,overlay]{textpos}
\usepackage{amsmath}
\usepackage{enumitem}
\usepackage{algorithm}
\usepackage{algorithmic}
\usepackage{booktabs}
\usepackage{tcolorbox}
\newtcolorbox{definition}{
  colback=gray!5,
  colframe=black,
  fonttitle=\bfseries,
  boxrule=0.5pt,
  arc=2pt,
  left=6pt,
  right=6pt,
  top=6pt,
  bottom=6pt
}
\newenvironment{boxedquote}
  {\begin{tcolorbox}[colback=gray!5,colframe=gray!20]}
  {\end{tcolorbox}}

\title{Tree of Thoughts as a Classical Heuristic Search Problem: Formal Foundations and Design Patterns}
\author {
    Guni Sharon
}
\affiliations{
    Texas A\&M University\\
    guni@tamu.edu
}

\begin{document}

\begin{textblock*}{0.9\textwidth}(0.5cm,0.5cm)
\small\itshape
This paper is an extended version of the paper published in the Proceedings of the Nineteenth International Symposium on Combinatorial Search (SoCS 2026).
\end{textblock*}
\maketitle

\begin{abstract} Large Language Models (LLMs) have demonstrated remarkable reasoning capabilities, yet their standard generation process---auto-regressive token prediction---is inherently myopic and prone to cascading errors. To address this, the \textit{Tree-of-Thoughts} (ToT) framework creates a search space over intermediate reasoning steps, allowing search models to explore, look ahead, and backtrack. However, current ToT research remains fragmented across \textit{Natural Language Processing} and \textit{Automated Planning} communities, often using inconsistent terminology and ad-hoc implementations. Consequently, we synthesize the ToT landscape through a unified taxonomy based on \textit{classical heuristic search terminology}. We map LLM-based reasoning to classical search components: state representation (granularity of thoughts), successor generation (prompting operators), and heuristic evaluation (self-assessment of progress). We analyze existing work within the context of our taxonomy and identify emerging design patterns: systematic search (Best-First Search) for shallow, deterministic tasks and lookahead-heavy strategies (DFS, MCTS) for deep multi-step reasoning. We conclude by identifying open algorithmic challenges at the 
intersection of heuristic search and LLM reasoning, and call 
on the heuristic search community to engage with this emerging domain. \end{abstract}

\section{Introduction}
Large Language Models (LLMs) have demonstrated remarkable capabilities in reasoning and problem-solving across a wide range of domains, from creative writing to mathematical proofs. Central to this success is the \textit{Chain of Thought} (CoT) prompting paradigm~\cite{wei2022chain}, which encourages models to generate intermediate reasoning steps before arriving at a final answer. However, CoT is fundamentally linear: once a reasoning step is generated, the model commits to it, often propagating early errors into final failures. These limitations are particularly pronounced in planning domains; \citet{valmeekam2023planning} conclude that ``LLMs' ability to generate executable plans autonomously is rather limited.''

To address this, recent work introduced the \textit{Tree-of-Thoughts} (ToT) framework~\cite{yao2023tree}, which generalizes CoT by maintaining multiple active reasoning paths and allowing lookahead and backtracking, effectively defining a tree search problem over a semantic state space. While ToT demonstrated significant accuracy improvements over CoT on various reasoning tasks, it lacks a unified taxonomy and consistent terminology. This lack of consistency limits (i) meaningful comparison between methods, (ii) understanding of design trade-offs, and (iii) transfer of established techniques from the heuristic search literature.

Addressing this gap, we propose a unified taxonomy mapping ToT to classical search components:
\begin{itemize}
\item \textbf{State Representation:} a sequence of ``thoughts''.
\item \textbf{Successor Generation:} LLMs as generative successor functions.
\item \textbf{Heuristic Evaluation:} Leveraging LLMs to estimate progress toward a goal.
\end{itemize}
By grounding ToT in classical search terminology, we enable systematic transfer of established search techniques to a domain with novel characteristics:\
\textbf{(1) Stochastic partial expansion:} Repeated expansion 
of the same state yields different successors;\
\textbf{(2) Implicit heuristics:} State evaluation emerges from learned patterns rather than domain-specific functions;\
\textbf{(3) High-dimensional linguistic state spaces:} States are text sequences, requiring new approaches to symmetry detection and state abstraction;\
and \textbf{(4) Hybrid objectives:} optimization targets combine solution correctness with qualitative measures such as fluency or diversity.\
These characteristics present novel algorithmic challenges that the search community is uniquely positioned to address. Beyond the algorithmic interest, LLM-based reasoning sits at the frontier of AI, with direct impact on automated decision-making, scientific discovery, and software engineering. Consequently, advances in ToT search algorithms may directly improve the reliability and effectiveness of LLM-based planning and reasoning systems.

\section{Preliminaries}\label{sec:prelim}
We briefly review \textit{Large Language Models} (LLMs) and structured prompting techniques, introduce the ToT framework~\cite{yao2023tree}, and establish the heuristic search concepts that will be applied throughout the paper.

\subsection{LLMs and Autoregressive Generation}
LLMs are autoregressive models~\cite{vaswani2017attention} that generate text by iteratively predicting the next token conditioned on a prefix sequence.

\begin{definition}

\paragraph{Tokens.}
A \emph{token} is the atomic unit of text processed by a language model. Tokens typically correspond to subword units produced by a fixed tokenizer (e.g., Byte Pair Encoding~\cite{sennrich2016bpe}) and may represent words, word fragments, punctuation, or special symbols. Higher-level textual constructs, such as sentences or ``thoughts’’, are therefore represented as sequences of tokens.
\end{definition}

Given an input prompt $x$ and a generated prefix $y_{1:t-1}$,\footnote{We use the slice notation $y_{i:j}$ to denote the sequence $y_i,y_{i+1}, \ldots, y_j$ throughout the paper.} the model defines a distribution
$p(y_t \mid x, y_{1:t-1})$,
which represents the probability of the next token $y_t$ conditioned on the prompt and previously generated tokens. 
During generation, the next token is drawn from this distribution according to a chosen \emph{decoding strategy}. Repeating this process produces a token sequence.

\begin{definition}
\paragraph{Decoding Strategies.}
A \textit{decoding strategy} specifies how the next output token is selected from the model's conditional distribution at each generation step. Common strategies include greedy decoding~\cite{graves2012sequence}, beam search~\cite{freitag2017beam}, top-$k$ sampling~\cite{fan2018hierarchical}, and nucleus (top-$p$) sampling~\cite{holtzman2019curious}. These strategies trade off determinism, diversity, and computational cost, and implicitly define a local search policy over the token-level decision space.
\end{definition}

Standard decoding strategies operate at the token level---at 
each step, one or more candidate tokens are selected and 
appended to growing sequences. Even strategies that maintain 
multiple candidates, such as beam search, evaluate alternatives 
based on token-level likelihood rather than semantic reasoning. Consequently, they have limited capacity to pause at 
intermediate reasoning steps, evaluate progress toward a goal, 
or backtrack at the level of meaningful semantic reasoning units 
(``thoughts'').

\subsection{Prompting and Intermediate Reasoning}
Prompting strategies~\cite{liu2023pre} are methods of structuring the input prompt $x$ to shape the LLM's conditional token distribution $p(y_t \mid x, y_{1
:t-1})$. They commonly include specific instructions, examples, or output format specifications with the objective of directing the LLM toward accurate and task-appropriate output distributions, from which a given decoding strategy will sample the output.
A prominent example is \emph{Chain-of-Thought} (CoT) prompting~\cite{wei2022chain,kojima2022large}, which encourages the model to generate explicit intermediate reasoning steps before producing a final answer. CoT has been shown to improve performance on arithmetic, logical, and symbolic reasoning tasks.

However, CoT is fundamentally linear and non-backtracking: the model commits to a single reasoning trajectory in a greedy fashion. Within the framework of combinatorial search, this corresponds to exploring a single reasoning path where errors in early search nodes (intermediate steps) inevitably lead to suboptimal or incorrect leaf nodes.
This limitation motivates approaches that explicitly branch, evaluate, and explore multiple intermediate reasoning paths.

\subsection{Tree of Thoughts (ToT)}
ToT~\cite{yao2023tree} extends Chain-of-Thought by framing reasoning as a structured search process over intermediate textual steps, referred to as \emph{thoughts}. 

\begin{definition}
\paragraph{Thoughts.}
A \textit{thought} is a contiguous sequence of tokens generated by a language model under a specified decoding strategy until a designated stopping condition is reached (e.g., a delimiter token, newline, or length threshold). Each node in the ToT is associated with a candidate thought generated from its parent node. The segmentation of text into thoughts is not uniquely defined and may be specified explicitly via prompt structure or stopping conditions, or implicitly induced by the language model during generation. For example, in planning domains~\cite{valmeekam2023planning}, a thought may correspond to a single action (e.g., \texttt{"pick block A and place on block B"}) or to a sequence of related actions (a macro-operator).
\end{definition}

Instead of generating a single linear sequence of thoughts as in CoT prompting, ToT maintains multiple candidate reasoning paths organized in a tree structure. At each step, the model proposes multiple candidate thoughts, evaluates them, and selectively expands promising branches.
Conceptually, ToT separates the reasoning process into four components:
\begin{enumerate}
    \item A \emph{proposal mechanism} that generates candidate next thoughts from a given sequence of thoughts.
    \item An \emph{evaluation mechanism} that estimates the quality or promise of partial solutions (sequences of thoughts). 
    \item A \emph{goal test mechanism} that determines whether a given sequence of thoughts constitutes a valid solution. Note that some prior ToT formulations implicitly conflate evaluation and goal testing; we separate them for clarity and consistency with classical search formulations.
    \item A \emph{search strategy} that determines how the ToT is explored (e.g., depth-first, breadth-first, best-first).
\end{enumerate}

This decomposition allows ToT to incorporate classical search algorithms while operating over a high-dimensional, linguistically-defined state space.

Note that the ToT literature~\cite{yao2023tree} often uses the term \textit{intermediate reasoning step} inherited from Chain-of-Thought prompting~\cite{wei2022chain}. While commonly used, this term is ambiguous in the context of heuristic search---it may refer to an atomic reasoning unit (a thought), a partial solution (reasoning path), or a ToT node. Consequently, for precision in our formalization, we avoid this term and instead use \textit{thought} to denote the atomic unit of reasoning (as defined above), \textit{state} to denote a sequence of thoughts, and \textit{node} to denote a ToT element containing a state and associated metadata (e.g., 
$g$ and $h$ values). Note that a state here is represented as a sequence of thoughts rather than a world configuration. This is because a sequence of thoughts implicitly defines a world configuration---namely, the one resulting from applying the sequence of thoughts to the initial configuration---while also serving as the context required by the LLM for next-thought generation. The formal details of the state space are given in Section~\ref{sec:state_space}.

\subsection{Classical Search Terminology}
A search problem is defined over a directed, weighted \textit{state-space graph} $G = (S, E)$, where $S$ is a set of states (vertices) and $E \subseteq S \times S$ is a set of directed edges. The state-space graph is often too large to store explicitly. In such cases, a search problem is often defined implicitly via a tuple $(s_0, \mathcal{O}, \mathcal{G})$, 
where $s_0 \in S$ is the initial state, 
$\mathcal{O}: S \to 2^S$ is the successor function, 
and $\mathcal{G}: S \to \{\text{True}, \text{False}\}$ is 
the goal test.

A search problem induces a \emph{search tree} rooted at $s_0$, where each layer is obtained by applying $\mathcal{O}$ to all states in the previous layer. Each branch in the search tree corresponds to a unique path in $G$. Following notation by~\citet{russell2020artificial}, a search tree node $n$ is defined by:
\begin{itemize}
    \item $n.s$: the state in $G$ to which the node corresponds.
    \item $n.\text{parent}$: the predecessor node that generated $n$.
    \item $n.g$: the path cost (sum of edge weights) from $s_0$ to $n.s$.
\end{itemize}
For informed search~\cite{pearl1983heuristics}, a heuristic value $n.h$---an estimate of the cost-to-go from $n.s$ to a goal state---is also defined. Informed search algorithms typically use an evaluation function $n.f$ that combines $g$ and $h$ values (e.g., $f = g + h$ in A$^*$) to guide node expansion order. Classical properties such as admissibility and consistency of the heuristic provide optimality and completeness guarantees for algorithms like A$^*$.
%(e.g., $f = g + h$ in A$^*$~\cite{hart1968formal})

In contrast to traditional search domains, the ToT state space presents several challenges: the successor function is often stochastic, and both $g$-values and $h$-values are not naturally defined. Nevertheless, adapting heuristic search algorithms to ToT can provide principled approaches for balancing exploration, exploitation, and computational cost in LLM reasoning.

\subsection{Planning and Search with Language Models}

Prior work on the intersection of LLMs and planning presents 
two competing paradigms. The \textbf{first} treats LLMs as 
\emph{end-to-end planners}, directly prompting them to 
generate complete plans~\cite{huang2022language}. This 
approach is fundamentally limited: LLMs fail to reliably 
produce valid plans even in simple domains, as autoregressive 
generation provides no guarantee of precondition satisfaction 
or goal reachability~\cite{valmeekam2023planning,kambhampati2024llms}.

The \textbf{second} paradigm, exemplified by 
LLM+P~\cite{liu2023llmp}, uses LLMs solely as a 
\emph{translation layer}: the LLM converts a natural language 
problem description into a PDDL specification, which is 
handed off to a classical planner. While effective, this 
approach requires well-structured domains admitting clean 
PDDL representations and forgoes the generative flexibility 
of LLMs for open-ended tasks.

ToT-style methods occupy a middle ground. Rather than treating the LLM as a standalone planner or a mere translator, they introduce a separation of responsibilities between two distinct entities: a \emph{search agent} and the LLM. The search agent manages and explores the ToT---maintaining the search frontier, selecting nodes for expansion, and tracking search progress---while the LLM serves as a stateless component invoked at specific interaction points: generating candidate next thoughts from a given state and possibly estimating the promise of partial solutions. The search agent thus retains full control over the global search strategy, while the LLM contributes its generative and evaluative capabilities locally~\cite{hao2023reasoning,yao2023tree}. A concrete specification of the prompt-based interfaces through which the search agent communicates with the LLM---thought generation, heuristic evaluation, goal testing, and state validation---is provided in Appendix~\ref{sec:interface}.

This separation of responsibilities reveals a significant gap in current ToT research. The planning and heuristic search communities have developed decades of algorithmic machinery---admissible and consistent heuristics, pruning strategies, anytime algorithms, and learned cost-to-go functions~\cite{ferber2020nn,ferber2022bootstrapping,agostinelli2019rubik}
---that current ToT implementations largely ignore. 

We aim to bridge this gap by formalizing ToT as a classical search problem, clarifying the design space for successor generation, cost assignment, and heuristic definitions, and demonstrating how different design choices give rise to distinct search behaviors and algorithmic trade-offs.

\paragraph{Iterative Repair.}
ToT is a constructive search process where solutions are 
built incrementally. Alternative paradigms such as 
\emph{Iterative Repair} or \emph{Local Search} take a 
different approach: a complete solution is generated 
first and refined through iterative LLM calls. 
Prominent examples include 
Reflexion~\cite{shinn2024reflexion} and 
Self-Refine~\cite{madaan2024self}. These are discussed 
in Appendix~\ref{sec:alternatives}, along with their 
trade-offs compared to ToT-based approaches.

\section{Search Tree Formalization for ToT}\label{sec:tot}
We formalize ToT as a heuristic search problem defined by a state space $S$, successor function $\mathcal{O}$, cost function ($g$-value), heuristic function ($h$-value), and goal test $\mathcal{G}$.

A concrete instantiation of all components defined in this section is 
provided in Appendix~\ref{sec:case_study} for readers who prefer a running example alongside the formalism.

\subsection{State Space $S$}\label{sec:state_space}
A state $s \in S$ is defined as a sequence $s = [z_0, z_1, z_2, \dots, z_t]$, where $z_0$ is the \emph{problem prompt} and each $z_i$ for $i > 0$ is a thought. The problem prompt is a fixed token sequence provided by the user that encodes the problem description, any relevant context, the initial world configuration, and the goal conditions.
The initial state is $s_0 = [z_0]$, a singleton sequence containing only the problem prompt. The state space $S$ is the set of all reachable thought sequences.

Formally, a state is represented as a full sequence of thoughts rather than a single thought. This is because the sequence implicitly defines a unique world configuration---the one resulting from applying all thoughts in order to the initial world configuration---and therefore carries the same information as a classical world state, albeit implicitly. A more memory-efficient implementation could store only the single last thought introduced at each ToT node, reconstructing the full state when needed by following backpointers to the root and concatenating thoughts along the path. However, current work typically stores full states per ToT node, as it simplifies implementation, and state memory complexity is less of a concern since the ToT's memory footprint is generally dominated by the LLM itself. In line with common ToT conventions, we assume storing full states per ToT node. 

Note that while each thought sequence is unique by construction
(making the ToT a tree over thought-sequence states), distinct
sequences may induce the \emph{same} world configuration. Consequently,  classical graph-search 
optimization techniques (e.g., duplicate detection, symmetry breaking) 
may become applicable if the implicit world-state is 
recoverable.
%This is the case, for example, in domains admitting a state to world-configuration mapping (e.g., via external environment validation, see \ref{goal_test:external} in Section~\ref{sec:goal_test}) , but an open challenge for the general case (see OC3 in Section~\ref{sec:challenges}). %1234

\subsection{Successor Function}\label{sec:successor_function}
For any state $s=[z_0, z_1, \dots, z_t]$, the successor function $\mathcal{O}(s)$ returns a set of successor states of the form $s' = s + [z_{t+1}]$, where $z_{t+1}$ is a proposed next thought appended to the sequence $s$ and the `+' operator is sequence concatenation. In principle, the number of distinct next thoughts is exponential in the thought length (approximately $|V|^{\ell}$, where $V$ is the vocabulary size and $\ell$ is the thought length in tokens). Generating all potential next thoughts is therefore intractable in many cases.
To address this, ToT employs \emph{proposal mechanisms}~\cite{yao2023tree} that generate a small subset of candidate thoughts. This is akin to Partial Expansion of search nodes~\cite{Goldenberg2014}, where only a subset of successors are generated and added to the search frontier. We characterize proposal mechanisms along two orthogonal dimensions: \textit{Sampling Strategy} and \textit{Structural Constraints}.

\paragraph{Sampling Strategy.} Defines how candidate thoughts are generated from the LLM:
\begin{enumerate}[label=\textbf{S\arabic*.},leftmargin=*]
    \item \textbf{Independent Sampling.} \label{proposal_mechanism:sampling}
    Generate each thought independently by sampling from the LLM's output distribution using a token-level decoding strategy (e.g., top-$k$ sampling, nucleus/top-$p$ sampling). This process is repeated $b$ times to produce $b$ candidate successor thoughts. The randomness at the token level induces diversity in the generated thoughts.

    \item \textbf{Explicit Diversity Sampling.} \label{proposal_mechanism:diversity}
    Extend independent sampling with explicit diversity mechanisms that bias generation to encourage semantic or lexical diversity among candidates.  The goal here is to reduce redundancy and improve exploration coverage of alternative reasoning paths. Common approaches include penalizing tokens that would produce $n$-gram overlap with previously generated thoughts~\cite{li2016diversity}, enforcing embedding-based dissimilarity thresholds~\cite{reimers2019sentence}, or using diverse beam search variants~\cite{vijayakumar2018diverse}. 
    
    \item \textbf{LLM-Enumerated Proposals.} \label{proposal_mechanism:enumerated}
    Prompt the LLM to explicitly generate a list of $b$ distinct candidate thoughts in a single forward pass. In practice, this is implemented via prompts such as \texttt{"Generate 5 different options for the next step"}, where the LLM response is parsed into 5 thoughts. This leverages the model's ability to produce several alternatives in parallel while maintaining coherence with the current state~\cite{yao2023tree}. However, it is particularly susceptible to the `mode-seeking' nature of aligned LLMs, which tend to collapse toward a few high-probability reasoning paths~\cite{kirk2024understanding}, often resulting in lower semantic diversity than independent sampling.%,gu2024minillm}.
\end{enumerate}

\paragraph{Structural Constraints.} Optionally imposed to further restrict the space of valid thoughts:
\begin{enumerate}[label=\textbf{C\arabic*.},leftmargin=*]
   \item \textbf{Domain-Specific Constraints.} \label{constraint:domain}
    Restrict thoughts to valid elements of a domain's action schema~\cite{hao2023reasoning}, such as legal moves in a game or valid operators in a planning domain. Unlike grammar-based constraints (\ref{constraint:grammar}), which enforce adherence to a formal language grammar, domain-specific constraints restrict generation to a defined set of domain actions, enforced either via the LLM prompt (e.g., \texttt{"Suggest a valid move from: \{move list\}"}) or by the search agent filtering out thoughts that do not appear in the provided move list post-generation. Note that whether a valid domain action is also contextually applicable given the current world configuration is a separate concern, addressed by semantic constraints (\ref{constraint:semantic}).

    \item \textbf{Grammar-Based Constraints.} \label{constraint:grammar}
    Enforce adherence to a formal grammar or structured format (e.g., arithmetic expressions, logical formulas, or formal languages such as PDDL or SQL). This can be implemented via constrained LLM decoding, where the decoder is restricted to only produce tokens that yield a valid parse according to the grammar (e.g., ensuring generated SQL queries conform to valid SQL syntax~\cite{scholak2021picard}), or by the search agent filtering out thoughts that fail to parse post-generation.
    
    \item \textbf{Length Constraints.} \label{constraint:length}
    Bound the token length of generated thoughts, either through prompt instructions or explicit truncation, to control computational cost and maintain thought granularity.

    \item \textbf{Semantic Constraints.} \label{constraint:semantic}
    Filter thoughts based on contextual validity using external verifiers, simulators, or model-based checkers. Unlike domain-specific constraints, which check the syntactic validity of a thought in isolation, semantic constraints are state-dependent---they assess whether a thought is valid given the current world configuration induced by the preceding thoughts. For example, a syntactically valid planning action may still violate preconditions in the current state and should therefore be rejected. This filtering is performed exclusively by the search agent, which invokes an external validator after thought generation and discards any thoughts that fail the contextual validity check before adding them to the search frontier.
\end{enumerate}

Importantly, these dimensions (sampling strategies and structural constraints) are \emph{orthogonal and composable}: structural constraints (\ref{constraint:domain}--\ref{constraint:semantic}) can be combined with any sampling strategy (\ref{proposal_mechanism:sampling}--\ref{proposal_mechanism:enumerated}), and multiple constraints can be applied simultaneously. 
%For instance, one might use independent sampling (\ref{proposal_mechanism:sampling}) with both domain-specific constraints (\ref{constraint:domain}) restricting thoughts to valid Blocksworld operators, (enforced via the LLM prompt, e.g., \texttt{"Generate a next action using only the operators: pick and place"}), and length constraint (\ref{constraint:length}) limiting thoughts to 10 tokens, enforced at the decoding level by setting a maximum token generation limit in the LLM decoding configuration.%1234

The choice of proposal mechanism directly determines the branching factor, diversity, and computational cost, and implicitly defines which regions of the thought space are explored.

\subsection{Pruning at the ToT Node Level }\label{sec:pruning}
Beyond thought-level filtering, which bounds the branching factor of the ToT, 
node-level pruning (applied post-generation) is commonly used to further reduce 
the ToT size and enable practical runtimes for tree search algorithms. Node-level 
pruning is typically guided by a node priority value. To conform with classical search notation, we denote this priority as the node's $f$-value and adopt the decomposition $n.f = n.g + n.h$, where potential instantiations of $g$ and $h$ are discussed in Sections~\ref{sec:cost} and~\ref{sec:heuristic}. However, unlike classical search where $n.g$ represents actual path cost and $n.h$ estimates cost-to-go, neither property is guaranteed in ToT settings: $g$ may be uninformative and $h$ often represents success likelihood rather than cost-to-go. Consequently, $f$-values in ToT should be interpreted as node ranking functions that guide search toward promising regions, rather than as estimates of the optimal path cost. As such, pruning nodes based on their $f$-value can effectively bias exploration toward high-quality solutions.
Common pruning strategies for ToT nodes  include:
\begin{enumerate}[label=\textbf{P\arabic*.},leftmargin=*]
    \item \textbf{Beam Pruning.}~\cite{yao2023tree} \label{pruning:beam}
    At each depth layer, retain only the top-$k$ nodes ranked by $f$-value across the entire layer, pruning all others. This corresponds to fixed-width best-first expansion without backtracking: once a node is pruned from the beam, it cannot be reconsidered in later expansions. Beam pruning results in linear growth of the search tree, with a total of $O(k \cdot d)$ nodes across all layers up to depth $d$. Recent beam pruning approaches~\cite{elbaz2025portfolio} incorporate diversity objectives directly into the retention criterion, complementing diversity at the generation level (\ref{proposal_mechanism:diversity}).
    
    \item \textbf{Local Branching Pruning.} \label{pruning:local_branching}
    At each node expansion, retain only the top-$b$ successors ranked by $f$-value, pruning the rest. Unlike beam pruning, which operates globally across a depth layer, local branching pruning operates on a per-node basis. When $b > 1$, this results in exponential tree growth ($O(b^d)$ nodes at depth $d$), but with a reduced branching factor compared to retaining all successors.

    \item \textbf{Local Threshold Pruning.} \label{pruning:local_threshold}
    Prune any node with $n.f > \texttt{threshold}$. Unlike local branching pruning, which retains a fixed number of successors regardless of quality, threshold pruning retains all nodes below the threshold. This preserves multiple promising branches when they exist---beneficial for solution diversity---but cannot bound the branching factor and may result in no pruning when $f$-values are poorly calibrated.
\end{enumerate}

While necessary for tractable search, pruning methods may invalidate optimality guarantees. Nodes pruned early may lie on optimal solution paths, and the lack of re-expansion mechanisms prevents recovery from such errors.

Additionally, unlike classical successor functions, ToT proposal mechanisms are typically stochastic: repeated expansions of the same state may yield different successors. This stochasticity motivates the use of MCTS algorithms~\cite{browne2012survey}, which explicitly account for sampling variance by repeatedly resampling successors to balance exploration and exploitation.%1234

\subsection{Cost Function ($g$-value)}\label{sec:cost}
In classical search, $n.g$ represents the cost of the path from the initial state $s_0$ to state $n.s$. In ToT settings, however, the notion of ``cost'' for a sequence of thoughts is not universally established. Unlike classical domains where edge costs are typically given, ToT cost functions must often be inferred from general task requirements. Common cost function instantiations include:

\begin{enumerate}[label=\textbf{G\arabic*.},leftmargin=*]
    \item \textbf{Uniform Cost.} \label{cost:depth}
    $n.g$ is defined as the depth of the node in the ToT, equivalent to the number of thoughts in the sequence. For example, in the ``Game of 24'' task (see D1. in Appendix~\ref{sec:benchmarks}), each thought corresponds to a single arithmetic operation, and minimizing depth corresponds to finding solutions with fewer operations.
    
    \item \textbf{No Cost.} \label{cost:null}
    In tasks where no natural notion of incremental cost exists---such as creative writing---all edges in the search tree may be assigned zero cost: $n.g = 0$ for all $n$.
    
    \item \textbf{Negative Log-Likelihood (NLL) Cost.} \label{cost:nll}
    Define $n.g$ as the accumulated negative log-likelihood of the thought sequence in $n.s$:
    \[
    g([z_0, z_1, \ldots, z_t]) = -\sum_{i=1}^{t} \log p(z_i \mid z_0, \ldots, z_{i-1}),
    \]
    where $p(z_i \mid z_0, \ldots, z_{i-1})$ is the probability assigned by the LLM to thought $z_i$ given the preceding context, and the probability of each thought is computed as the product of the probabilities of its constituent tokens. This cost function naturally emerges when converting from a probabilistic framework to a cost-based framework, as the negative log transformation converts multiplicative probabilities into additive costs~\cite{meister2020beam}. Under this cost model, search favors paths through high-probability thought sequences. Note that unnormalized NLL costs inherently penalize longer reasoning chains, potentially biasing the search toward shorter, shallower reasoning paths. Consequently, it is advisable to apply length normalization~\cite{Wu2016GooglesNM}.
\end{enumerate}

The choice of cost function significantly impacts search behavior: it defines what constitutes an ``optimal'' solution and determines how the search algorithm trades off solution quality against path length.

\subsection{Heuristic Function ($h$-value)}\label{sec:heuristic}
In classical search, $h(s)$ estimates the cost-to-go from state $s$ to the nearest goal state satisfying $\mathcal{G}$. In ToT settings, $h(s)$ is challenging to formalize: goals may be underspecified (e.g., \texttt{"write a creative story"}) and distance to a goal is not naturally quantifiable in linguistic spaces.

A key distinction from classical search is that common ToT implementations~\cite{yao2023tree,shinn2024reflexion} define $h(s)$ as a \emph{success likelihood}---a score or probability indicating how promising a partial solution appears---rather than a cost-to-go estimate. We denote such heuristics as $h_{\text{success}}(s)$; nodes are ranked by maximizing $h_{\text{success}}$ rather than minimizing it. Since $h_{\text{success}}$ and $g$ have incompatible scales and directions of preference, they cannot be directly summed to form a meaningful $f$-value. To align with the classical cost-minimization framework, $h_{\text{success}}(s)$ can be converted to a cost-based heuristic, denoted $h_{\text{cost}}$, via inversion. For example, assuming $h_{\text{success}} \in [0,1]$:
\begin{itemize}
\item $h_{\text{cost}}(s) = (1 / h_{\text{success}}(s))-1$   

\item $h_{\text{cost}}(s) = -\log(h_{\text{success}}(s))$ 
\end{itemize}
The logarithmic transformation is particularly relevant when $h_{\text{success}}$ represents a probability, as it converts multiplicative probability relationships into additive costs, consistent with the NLL cost (\ref{cost:nll}). In the remainder of this section, heuristic functions are described as they appear in the ToT literature (typically success-based). 

\begin{enumerate}[label=\textbf{H\arabic*.},leftmargin=*]
    
    \item \textbf{Learned Evaluation} (type $h_{\text{success}}$). \label{h:value}
    The search agent invokes a function approximator to estimate the promise of a state. The approximator may be the LLM used for generation (e.g., prompted to rate a partial solution on a scale from 0 to 10)~\cite{yao2023tree}, a separate model acting as a discriminator~\cite{qi2024rstar}, a dedicated \textit{process reward model}~\cite{zhang2025,still1_2024}, a learned $Q$-value model~\cite{wang2024qstar}, or a learned cost-to-go function of the kind developed in classical planning~\cite{ferber2020nn,agostinelli2019rubik}. The estimate may be elicited as (i) a \emph{scalar score}; (ii) a \emph{categorical judgment} mapped to a numeric value (e.g., \textit{sure/maybe/impossible} $\rightarrow 2/1/0$)~\cite{yao2023tree}, with the binary case reducing to a validity filter; or (iii) a \emph{comparative judgment} over several sibling candidates~\cite{yao2023tree}. Note that (iii) yields only an ordering over the compared set, not a value on a common scale, and therefore cannot be summed with $g$ or used to rank nodes from different parents. What unites these instantiations, and separates them from \ref{h:external}, is that the value is fitted to data rather than computed, so admissibility does not follow from construction.

    \item \textbf{Thought Probability} (type $h_{\text{success}}$). \label{h:probability}
    Define $h(s)$ as the joint probability of the thought sequence in $s$~\cite{pendurkar2025policyguided}. Formally, for a state $s = [z_0, z_1, \ldots, z_t]$, 
    %\vspace{-1mm}
    $$h(s)=p(s) = \prod_{i=1}^{t} p(z_i \mid z_0, \ldots, z_{i-1})$$ 
    %\vspace{-1mm}
    Higher $h(s)$ indicates greater model confidence in the thought sequence. H2 rests on the assumption that $p(\text{success} \mid s) \approx p(s)$, i.e., that the probabilities of thought sequences approximate their likelihood of reaching a solution. Under a negative log transformation, H2 is functionally equivalent to NLL Cost (\ref{cost:nll})---both induce the same node ranking---and consequently using them jointly is redundant. Note that in~\cite{pendurkar2025policyguided}, $h(s)$ is used as part of a ratio cost function $\text{cost}(s) = g(s) / h(s)$ rather than the additive $n.f = n.g + n.h$ decomposition and $g(s)$ is set as \ref{cost:depth} rather than by mechanism: \ref{h:probability} reads the generating policy's own likelihood, whereas \ref{h:value} is a separately elicited estimate of promise. Both are produced by approximators.

    \item \textbf{External Functions} (potentially $h_{\text{cost}}$). \label{h:external}
    The search agent invokes a domain-specific external tool---such as a simulator, code interpreter, or formal verifier---to evaluate state quality~\cite{valmeekam2023planning}. Unlike Scalar Value Estimation (\ref{h:value}), the $h$-value is grounded in execution or symbolic reasoning rather than model-based approximation, providing more reliable feedback at the cost of requiring domain-specific tooling. Interestingly, recent work~\cite{correa2025classical} demonstrated that LLMs can independently generate such external heuristic functions.
\end{enumerate}

These evaluation methods serve as proxies for cost-to-go but differ
fundamentally from classical heuristics. LLM-based $h$-values
(\ref{h:value}, \ref{h:probability}) are not derived from domain
knowledge or problem relaxations but from the model's learned
associations and pattern matching. Consequently: (1) they may incur significant computational overhead and can potentially benefit from lazy evaluation~\cite{dellin2016lazysp}; and (2) classical properties
such as admissibility and consistency generally do not apply.
However, several avenues exist for recovering formal guarantees in
structured domains. First, admissible closed-form heuristics (e.g., gap heuristics~\cite{helmert2010landmark}, pattern databases~\cite{culberson1998pattern}) could be combined with LLM-based estimates, for instance, using the closed-form value as a floor, for recovering (bounded) admissibility guarantees. Second, consistency can often be recovered via
pathmax~\cite{mero1984pathmax}. Third, recent work on heuristic approximation provides relevant
theoretical guarantees regarding
admissibility~\cite{futuhi2026learning,ferber2022bootstrapping}.
Finally, we note that some effective search algorithms~\cite{hoffmann2001ff,Orseau2018lts,browne2012survey} operate without admissibility
guarantees.

\subsection{Goal Test}\label{sec:goal_test}
The goal test $\mathcal{G}: S \to \{\text{True}, \text{False}\}$ determines whether a state constitutes a valid solution. 

\begin{enumerate}[label=\textbf{T\arabic*.},leftmargin=*]
    \item \textbf{Deterministic Verification.} \label{goal_test:deterministic}
    The search agent verifies solution correctness via direct programmatic computation on the final output, without requiring an external environment. For example, in the ``Game of 24'' task, the goal test evaluates whether the final arithmetic expression equals 24; in programming tasks, unit tests verify functional correctness~\cite{chen2021evaluating,austin2021program}.
    This approach provides objective, reproducible validation when solution correctness can be determined from the output alone.

    \item \textbf{LLM-Based Evaluation.} \label{goal_test:llm}
    The search agent uses an LLM (potentially the same one used for generating states) to judge whether the final thought sequence satisfies the task requirements~\cite{yao2023tree}. For example, in creative writing tasks, the LLM may assess whether a generated story meets specified constraints (e.g., genre, theme). Notably, in such linguistic domains, the goal test is often a threshold over a continuous quality score which may be inconsistent across evaluations.

    \item \textbf{External Environment Validation.} \label{goal_test:external}
    The search agent verifies solutions by executing them in an external environment, such as a world simulator or interactive execution environment. Unlike Deterministic Verification (\ref{goal_test:deterministic}), correctness here depends on the state of the environment rather than the output alone---for example, in Blocksworld, $\mathcal{G}(s)$ is often determined by executing the plan in a simulator and checking whether the resulting world state matches the goal configuration~\cite{valmeekam2023planning}.
\end{enumerate}

In practice, hybrid goal tests combining multiple verifications are also possible. For example, a programming task might use both unit test execution (\ref{goal_test:deterministic}) and LLM-based evaluation of code quality or style (\ref{goal_test:llm})~\cite{rozi2024codellama}.

\paragraph{State Validator.}
Goal tests can often be extended to serve as a \textit{state validator}---a function that determines whether a partial solution defined by $s$ is valid or 
executable~\cite{valmeekam2023planning}. 
For example, in the Game of 24 domain, a state validator would reject partial 
solutions that lead to division by zero or produce non-numeric results. 
State validators enable early pruning of invalid branches (dead-end detection), 
improving search efficiency by avoiding expansion of states that cannot possibly 
lead to valid solutions. State validators are typically applied as semantic constraints (\ref{constraint:semantic}) during successor generation.

\section{ToT Design Choices}\label{sec:design_choices}
A practical implementation of ToT must commit to concrete instantiations of the interdependent components described above: (i) the proposal mechanism defining the successor function $\mathcal{O}(s)$, possibly with structural constraints (Section~\ref{sec:successor_function}), (ii) node pruning strategies for tractable computation (Section~\ref{sec:pruning}), (iii) the path cost function $g(s)$ (Section~\ref{sec:cost}), (iv) the heuristic function $h(s)$ (Section~\ref{sec:heuristic}), (v) the goal test $\mathcal{G}$ (Section~\ref{sec:goal_test}), and finally, (vi) the search algorithm that is deployed to traverse the resulting ToT.
 
While these components are defined independently in the formal framework, in practice their choices are tightly coupled, and some combinations are more natural or effective than others. The task semantics often dictate whether cost-sensitive search is meaningful (i.e., whether minimizing $g$ is a relevant objective), whether heuristic guidance should be scalar-valued or comparative, and whether successor generation should prioritize diversity or likelihood. As a result, ToT implementations typically adopt coherent design patterns rather than arbitrary combinations of components.
 
\paragraph{Two trees.} Throughout this section we distinguish the \emph{problem graph}, whose vertices are world configurations and whose edges are domain actions, from the \emph{ToT}, whose vertices are thought sequences and whose edges are generated thoughts. The ToT is a sampled reasoning structure induced by the problem graph (Sections~\ref{sec:state_space} and~\ref{sec:successor_function}). Solution depth and branching factor are properties of the problem graph and are stated there; thought granularity and node count are properties of the ToT. Solution depth is therefore well defined only for domains admitting a formal specification, and where it is not, counting thoughts measures the chosen reasoning representation rather than an intrinsic property of the task. Granularity need not even be uniform within one tree: in~\cite{qi2024rstar} a single thought $z_{t+1}$ may span anything from one reasoning step to the entire remaining chain, so $t$ does not measure progress consistently across branches.
 
\subsection{Design Choices in Prior Work}\label{sec:prior_work}
 
We survey representative ToT implementations from the literature and analyze their design choices along the axes established in Section~\ref{sec:tot}. Table~\ref{tab:design_choices} summarizes these implementations. Since most of the surveyed works do not describe themselves in search-theoretic terms, the entries reflect our reading of the reported algorithms and, where available, of the released implementations.
 
\paragraph{Reading Table~\ref{tab:design_choices}.} Two caveats apply.
First, where an implementation applies no goal test during search, the search is assumed to terminate due to a depth limit or end-of-sequence condition, after which validity is checked on the terminal state. The \textbf{Goal Test} column records that validity check.
Second, one row is a deliberate boundary case. PICARD~\cite{scholak2021picard} constrains decoding at \emph{token} granularity rather than at the level of semantic reasoning units, and therefore sits at the lower limit of the thought abstraction (Section~\ref{sec:prelim}). We retain it because it is the clearest instance of grammar-based constraints (\ref{constraint:grammar}) applied inside a search loop, and mark it with $\dagger$.
 
\begin{table*}[t] \centering \resizebox{\textwidth}{!}{
\begin{tabular}{@{}lccccccc@{}} \toprule \textbf{Work} & \textbf{Domain} & \textbf{Proposal} & \textbf{Pruning} & \textbf{Search Strategy} & \textbf{$g$-value} & \textbf{$h$-value} & \textbf{Goal Test} \\
\midrule
\cite{zhang2025cumulative} & Logical Inference & \ref{proposal_mechanism:enumerated} + \ref{constraint:semantic} & validity filter & Greedy (DAG) & \ref{cost:null} & \ref{h:value} & \ref{goal_test:llm} \\
 
\cite{xie2023selfevaluation} & Math Reasoning & \ref{proposal_mechanism:sampling} & \ref{pruning:beam} & Beam & \ref{cost:nll} & \ref{h:value} & \ref{goal_test:deterministic}$^{r}$ \\
 
\cite{scholak2021picard}$^{\dagger}$ & Text-to-SQL & beam expansion + \ref{constraint:grammar} & \ref{pruning:beam} & Beam & \ref{cost:nll} & --- & --- \\
 
\cite{yao2023tree} & Game of 24 & \ref{proposal_mechanism:enumerated} + \ref{constraint:domain} & \ref{pruning:beam} & Beam & \ref{cost:null} & \ref{h:value} & \ref{goal_test:deterministic} \\
\cite{yao2023tree} & Creative Writing & \ref{proposal_mechanism:sampling} & \ref{pruning:beam} ($k{=}1$) & Beam & \ref{cost:null} & \ref{h:value}$^{c}$ & \ref{goal_test:llm} \\
\cite{yao2023tree} & Crosswords & \ref{proposal_mechanism:enumerated} + \ref{constraint:semantic} & \ref{pruning:local_threshold} & DFS & \ref{cost:null} & \ref{h:value} & \ref{goal_test:deterministic}\\
 
\cite{pendurkar2025policyguided} & Game of 24 & \ref{proposal_mechanism:sampling} + \ref{constraint:domain} & \ref{pruning:local_branching} & LTS & \ref{cost:depth} & \ref{h:probability} & \ref{goal_test:deterministic} \\
\cite{pendurkar2025policyguided} & Blocksworld & \ref{proposal_mechanism:sampling} + \ref{constraint:domain} & \ref{pruning:local_branching} & LTS & \ref{cost:depth} & \ref{h:probability} & \ref{goal_test:external} \\
 
\cite{still1_2024} & Math (competition) & \ref{proposal_mechanism:sampling} & --- & MCTS & \ref{cost:null} & \ref{h:value} & \ref{goal_test:deterministic}$^{r}$ \\
 
\cite{hao2023reasoning} & Blocksworld & enumerated + \ref{constraint:domain} + \ref{constraint:semantic} & --- & MCTS & \ref{cost:null} & \ref{h:value} + \ref{h:probability} & \ref{goal_test:deterministic}$^{w}$ \\
 
\cite{zhou2023language} & Programming
  & \ref{proposal_mechanism:sampling}+\ref{constraint:domain}+\ref{constraint:semantic}
  & --- & MCTS & \ref{cost:null} & \ref{h:value} & \ref{goal_test:deterministic} \\
 
\cite{qi2024rstar} & Math Reasoning & \ref{proposal_mechanism:sampling} + \ref{constraint:domain} & \ref{pruning:local_branching} & MCTS & \ref{cost:null} & \ref{h:value} & \ref{goal_test:deterministic}$^{r}$ \\
 
\bottomrule \end{tabular}
} \caption{Design choices in representative ToT implementations. Labels reference the taxonomy defined in Section~\ref{sec:tot}. \textbf{Beam} denotes level-by-level breadth-first beam search. \textbf{DFS} denotes greedy depth-first search~\cite{russell2020artificial}. \textbf{LTS} denotes Levin Tree Search~\cite{Orseau2018lts}. \textbf{Greedy (DAG)} denotes non-backtracking accumulation of verified propositions into a directed acyclic graph. Markers: $\dagger$~token-granularity boundary case; $^{c}$~comparative vote over sibling candidates rather than an independent score; $^{r}$~correctness established against a reference answer, so the check is not one the search agent could invoke; $^{w}$~goal conditions checked against a world state maintained by the LLM rather than an external simulator. Under level-synchronous expansion (Beam), all frontier nodes share a depth, so \ref{cost:depth} and \ref{cost:null} induce identical rankings; we report \ref{cost:null} where the implementation defines no explicit cost. A dash in the \textbf{Pruning} column indicates that no explicit pruning rule is reported; in the MCTS rows expansion is governed by the selection policy instead. Details regarding the benchmark domains and evaluation metrics are provided in Appendix~\ref{sec:evaluation}.} \label{tab:design_choices} \end{table*}
 
\paragraph{Design Patterns.} Table~\ref{tab:design_choices} exhibits two correlations between task structure and design choice. We state each with the rows that instantiate it. The surveyed set is not exhaustive, and we do not claim that these correlations are exclusive or causal.
\paragraph{(1)~~Solution depth and traversal order.}
Where solutions are shallow, as in Game of 24 at depth three, implementations keep several alternatives live and rank them against one another before carrying any forward: beam search ranks all successors of a level and retains the best~\cite{yao2023tree}, and LTS ranks the entire frontier by Levin cost~\cite{pendurkar2025policyguided}. Where solutions are longer, as in crosswords at up to ten steps, Blocksworld at two to six, and programming with no fixed length, implementations instead carry one path to a terminal state before comparing it against alternatives, whether by descending until a state is judged unpromising (DFS~\cite{yao2023tree}) or by completing a rollout and backpropagating its value (MCTS~\cite{hao2023reasoning,zhou2023language}).
\paragraph{Levin Tree Search spans both sides.} LTS~\cite{pendurkar2025policyguided} appears on both sides of this contrast and belongs exclusively to neither regime. It expands nodes in order of $d(n)/\pi(n)$, where $\pi(n)$ is the product of policy probabilities along the path to $n$, and the policy temperature interpolates between the two sides of the contrast above: as the temperature falls, LTS carries one path to a terminal before comparing it against alternatives, and as it rises, LTS ranks an entire layer before descending.\footnote{As the policy concentrates, $\pi \rightarrow 1$ along the greedy path and $\pi \rightarrow 0$ elsewhere, so expansion proceeds greedily depth-first. As the policy approaches uniform, $\pi(n) = b^{-d}$ and the priority $d\,b^{d}$ is constant within a level for uniform $b$ and increasing in $d$, so expansion proceeds breadth-first.} Where the surveyed implementations fix a traversal order by construction, LTS exposes it as a parameter.
 
\paragraph{(2)~~Open-ended tasks: LLM-based evaluation and goal testing.}
Creative writing~\cite{yao2023tree} admits no programmatic checker, and the corresponding implementation uses the LLM for both the heuristic and the goal test (\ref{h:value}, \ref{goal_test:llm}). Without a formal criterion, neither validity nor quality can be computed, so no other option is available. A consequence is that pruning discards branches that ranked lower rather than branches that are invalid, and is therefore unsound.
 
\paragraph{An absence: no computed heuristic.} No implementation surveyed here obtains a heuristic value by computation over an explicit model of the domain (\ref{h:external}). Every $h$-value in Table~\ref{tab:design_choices} is produced by a function approximator, including in the rows whose goal test the search agent can itself invoke, such as evaluating an expression or running a simulator. Among the surveyed implementations, grounded computation thus enters as a check on terminal states but never as an estimate of remaining distance. Since admissibility is what converts a heuristic into a guarantee, its absence removes the basis for bounded-suboptimal search and for any guarantee on solution quality (OC2). This holds equally of recent best-first methods that adopt classical algorithms wholesale. ToolChain$^*$~\cite{zhuang2024toolchain} and Q$^*$~\cite{wang2024qstar} both run A$^*$ over LLM reasoning states, but their $h$ is estimated rather than computed: from stored plans together with an LLM-generated guess at the remaining path length in the former, and from a learned $Q$-value model in the latter. Neither carries an admissibility guarantee, so the algorithm transfers without the guarantees that motivate it.

\section{Open Challenges and Research Directions}\label{sec:challenges}

The formalization presented in this work exposes a set of concrete algorithmic gaps---places where classical search theory makes precise predictions, but ToT implementations lack the machinery to act on them. We organize these gaps into three research directions.

\paragraph{OC1. Adapting Classical Algorithms.}
Current ToT implementations can benefit from, yet often 
overlook, a wealth of classical search techniques. Promising candidates include 
anytime and bounded-suboptimal algorithms (e.g., weighted 
$A^*$, $ARA^*$~\cite{likhachev2004ara}) for trading 
solution quality against computation time; pruning 
techniques such as partial-order 
reduction~\cite{Wehrle2012}, Partial 
Expansion~\cite{Goldenberg2014}, and symmetry 
breaking~\cite{domshlak2012symmetry} for reducing the 
effective branching factor; and lazy evaluation 
frameworks~\cite{dellin2016lazysp} for deferring expensive 
LLM calls. Transferring these techniques is not straightforward: suboptimality bounds assume meaningful 
$g$-values, while ToT costs are often zero 
(\ref{cost:null}) or length-biased (\ref{cost:nll}); and 
symmetry detection requires
mapping thought sequences to world configurations (see OC3).

\paragraph{OC2. Heuristic Design and Learning.}
Given their inherent output randomness, LLM-based 
heuristics (\ref{h:value}, \ref{h:probability}) largely 
lack admissibility guarantees. While recent 
work~\cite{futuhi2026learning,ferber2022bootstrapping} 
provides such guarantees in specific domains, generalizing 
them remains an open challenge. Improved heuristic 
approximation can feed a positive loop where successful search 
traces enable fine-tuning the LLM's heuristic estimation---e.g., using STaR-style~\cite{zelikman2022reasoner} or 
DPO~\cite{rafailov2023direct} methods---which in turn 
yields better search traces.

\paragraph{OC3. Novel Search Challenges and Theory.}
ToT introduces unique algorithmic challenges. First, \textit{Stochastic partial expansion}: repeated expansion of the same state yields different successors, violating determinism assumptions underlying most completeness proofs. While MCTS~\cite{browne2012survey} partially addresses this via resampling, convergence proofs assume $f$-value estimates stabilize with repeated sampling---an assumption that may be violated when LLM outputs are non-stationary. Second, \textit{Semantic state equivalence}: distinct ToT states (thought sequences) may encode identical world configurations. Reliable duplicate detection has significant pruning potential, yet it remains an open problem. Third, \textit{Token-budget search}: LLM inference cost scales with token count; establishing performance bounds for anytime and bounded-suboptimal algorithms under a token budget requires new theoretical machinery. Finally, progress on these challenges would benefit from standardized 
benchmarks annotated with search-theoretic properties 
(optimal solution depth, branching factor), enabling 
controlled comparison of search efficiency across methods 
(see Appendix~\ref{sec:evaluation}).

% \paragraph{OC4. Benchmarking and Evaluation.}
% Progress in ToT is impeded by the absence of standardized benchmarks with known search-theoretic properties---specifically, problem sets annotated with optimal solution depth and branching factor, enabling controlled comparison of search efficiency across methods. We advocate for benchmark construction that explicitly varies these parameters, analogous to parametric problem generators in classical planning competitions, alongside separate reporting of heuristic quality metrics and task-level success rates (see Appendix~\ref{sec:evaluation}).

\section{Conclusion}\label{sec:conclusion}
We presented a unified taxonomy of the Tree-of-Thoughts framework grounded in classical heuristic search, mapping LLM-based reasoning to classical search components: state representation, successor generation, cost evaluation, and heuristic guidance. While ToT implementations vary in their prompting techniques, they converge on a small set of fundamental search patterns tailored to domain structure: systematic search (BFS) for shallow deterministic tasks and lookahead-heavy strategies (DFS, MCTS) for deep reasoning. This perspective allows us to view current LLM agents not as distinct ``prompting tricks,'' but as specific instantiations of well-studied search algorithms.\\
\textbf{A Call to the Search Community.}
Our formalization reveals ToT as more than an NLP technique---it is a heuristic search problem in a fundamentally new domain. Current implementations typically employ basic BFS or DFS without leveraging decades of advances in search algorithms, heuristic design, and pruning techniques. Meanwhile, the unique characteristics of ToT---stochastic successor generation, learned heuristics, linguistic state spaces, and computational budgeting---present novel algorithmic challenges that the heuristic search community is uniquely equipped to address.
We hope this work serves as a foundation for systematic engagement between the heuristic search and LLM communities, enabling the transfer of algorithmic insights in both directions: proven search techniques can improve ToT implementations, while the novel characteristics of learned search spaces can inspire new theoretical and algorithmic advances in heuristic search.

\section*{Acknowledgment}
This paper benefited from the support of \textbf{Ariel Felner}, who  encouraged me to write this paper and offered valuable structural and editorial comments.
The research for this paper was supported in part by the NSF (IIS-2238979).

\bibliography{ref}

\appendix

\clearpage

\section{Search Agent-LLM Interface}\label{sec:interface}
The search agent interfaces with the LLM through prompt-based queries. Since the LLM is stateless, each query must include a complete thought sequence, i.e., $s = [z_0, z_1, \ldots, z_t]$, to provide necessary context. We identify four distinct interface types corresponding to different search components:

\paragraph{Interface 1: Thought Generation.} Used to implement the successor function $\mathcal{O}(s)$ (Section~\ref{sec:successor_function}).
\begin{itemize}
\item \textbf{Example Prompt:} $[z_0, z_1, \ldots, z_t]$ + \textit{"Suggest a next step"} + Structural Constraints (\ref{constraint:domain}--\ref{constraint:semantic})

\item \textbf{Response:} A candidate thought $z_{t+1}$ generated according to the designated sampling strategy (\ref{proposal_mechanism:sampling}--\ref{proposal_mechanism:enumerated}). Optionally, the search agent computes likelihood-based costs (\ref{cost:nll}) or probability-based heuristics (\ref{h:probability}) by extracting the token-level probabilities $p(z_{t+1} \mid z_0, \ldots, z_t)$ returned as metadata by the LLM API during generation.
\end{itemize}

\paragraph{Interface 2: LLM-Based Heuristic Evaluation.} Used to compute
$h$-values when the heuristic relies on LLM assessment (\ref{h:value}, \ref{h:probability}).
\begin{itemize}
\item \textbf{Example Prompt:} $[z_0, z_1, \ldots, z_t]$ + \textit{"Estimate the number of extra steps required to solve the problem"}.

\item \textbf{Response:} A numerical
$h$-value or categorical assessment that is mapped to a numerical value.
\end{itemize}

\paragraph{Interface 3: LLM-Based Goal Test.} Used to implement the goal test $\mathcal{G}(s)$ when relying on LLM judgment (\ref{goal_test:llm}).
\begin{itemize}
\item \textbf{Example Prompt:} $[z_0, z_1, \ldots, z_t]$
 + \textit{"Is this a valid solution?"}

\item \textbf{Response:} Boolean (\texttt{True}/\texttt{False}) or text that is parsed to extract a binary decision.
\end{itemize}

\paragraph{Interface 4: LLM-Based State Validator.}
Used to check partial solution validity, enabling semantic constraints (\ref{constraint:semantic}) and early pruning of invalid branches.
\begin{itemize}
\item \textbf{Prompt:} $[z_0, z_1, \ldots, z_t]$
+ \textit{"Is this a valid partial solution?"}

\item \textbf{Response:} Boolean (\texttt{True}/\texttt{False}) indicating whether the partial state is consistent and executable.
\end{itemize}

\paragraph{External Components.}
LLM interfaces \#2--4 can be substituted by external components. The search agent then requires separate interfaces to these external components, such as external heuristics (\ref{h:external}), deterministic goal tests (\ref{goal_test:deterministic}), or environment-based validation (\ref{goal_test:external}). Such external interfaces operate independently of the LLM and typically provide more reliable signals than LLM-based assessment, at the cost of requiring domain-specific tooling.

\paragraph{Search Agent Operations.}
Importantly, node-level pruning (Section~\ref{sec:pruning}) and search control decisions (e.g., node selection, frontier management) are handled exclusively within the search agent itself. The search agent maintains the search tree structure, tracks visited states, manages the open list (frontier), and orchestrates the overall search process according to a given algorithm (BFS, DFS, MCTS, etc.).

\paragraph{Cache Efficiency.}
Beyond algorithmic design, hardware realities influence the practical
viability of search strategies in ToT. In particular, LLM inference
relies on a key-value cache (KV-cache) that stores intermediate
attention computations. DFS-like traversal naturally extends this
cache incrementally, whereas BFS or MCTS require context switching
between branches, incurring recomputation costs. However, recent
systems-level advances such as prefix
caching~\cite{kwon2023vllm} enable efficient
KV-cache reuse for states sharing a common prefix---the typical case
in ToT, where all states share the root prompt $z_0$. As these
optimizations mature, the gap between DFS and alternative traversal
orders is expected to narrow, though the relative cache efficiency of
different search strategies remains a practical consideration for
current implementations.

\section{Search over Complete Solutions}
\label{sec:alternatives}

While ToT frames reasoning as a \emph{constructive search} problem---where the search tree is built by incrementally extending partial solutions---an alternative paradigm performs search over the space of \emph{complete} solutions. In classical search literature, this corresponds to \emph{Iterative Repair} or \emph{Local Search}.

Prominent examples include \emph{Reflexion}~\cite{shinn2024reflexion}, \emph{Self-Refine}~\cite{madaan2024self}, and recent work in generalized planning such as \emph{PG3}~\cite{silver2024generalized}. In these frameworks, a state $s$ represents a full candidate solution (e.g., a complete code snippet or a full essay). The successor function $\mathcal{O}(s)$ does not extend the solution, but rather \emph{revises} it based on feedback or self-critique:
\[
    s_{t+1} \leftarrow \text{Refine}(s_t, \text{Feedback}(s_t))
\]
This approach simplifies the search topology from a tree to a single trajectory (or a set of disconnected trajectories), effectively performing hill-climbing on the quality surface of the answer.

\paragraph{Trade-offs.} The choice between Constructive Search (ToT) and Iterative Repair involves balancing \textit{contextual visibility} against \textit{error recovery}:

\begin{itemize}
    \item \textbf{Iterative Repair} operates with lower (linear) search overhead and benefits from \emph{global context}---the ability to critique the solution as a complete whole. This is ideal for refinement tasks where the initial solution (a complete sequence of thoughts) provides a strong structural prior. However, it is highly susceptible to local optima: if the initial solution is fundamentally flawed, Iterative Repair often lacks the mechanisms to make large structural jumps or explore distinct alternatives.
    
    \item \textbf{Constructive Search (ToT)}, conversely, is limited by \emph{greedy lookahead}: it must commit to partial solutions without knowing if they lead to a valid full solution. However, by maintaining a frontier of multiple partial solutions, it retains the ability to \textit{backtrack} and explore alternative solutions, making it superior for precise, ``compositional'' domains where a single early error can invalidate subsequent solutions.
\end{itemize}

\section{Evaluation}\label{sec:evaluation}

ToT methods operate over high-dimensional, stochastic, and partially structured state spaces. As a result, their evaluation requires careful consideration of both the problem domains and the metrics used to assess performance. 
In this section, we first characterize the benchmark domains used in prior ToT research, organizing them by their structural properties and verification mechanisms. We then discuss common evaluation metrics, noting that effective ToT evaluation typically requires combining task-level performance measures with search-centered efficiency metrics, rather than relying on a single scalar measure.

\subsection{Benchmark Domains}\label{sec:benchmarks}

ToT methods have been evaluated across a diverse range of domains, each presenting distinct characteristics in terms of state space structure, solution depth, branching factor, and verification methods. We categorize representative benchmark domains based on their primary characteristics:

\paragraph{D1. ~~Constraint Satisfaction and Puzzle Solving.}
These domains feature well-defined goals, deterministic verification, and typically shallow solution depths:

\begin{itemize}
\item \textbf{Game of 24}~\cite{yao2023tree}: Given four numbers, find arithmetic operations that result in 24. Solutions are typically 3--4 steps deep, with high branching factor due to operation choice. Success is binary and deterministically verifiable via expression evaluation.

\item \textbf{Crosswords}~\cite{yao2023tree}: Fill in crossword puzzles given clues. Solutions require satisfying multiple intersecting constraints, with word validity checkable via dictionary lookup. The challenge lies in managing interdependencies between intersecting words.

\item \textbf{Logic Puzzles}~\cite{zhang2025cumulative}: Solve deductive reasoning problems (e.g., Sudoku, Einstein puzzles). These feature complex constraint networks and require systematic reasoning to eliminate possibilities. Solutions are deterministically verifiable against puzzle constraints.
\end{itemize}

\paragraph{D2. ~~Mathematical and Symbolic Reasoning.}
Domains requiring multi-step symbolic manipulation and arithmetic:

\begin{itemize}
\item \textbf{Math Reasoning}~\cite{xie2023selfevaluation,wei2022chain}: Solve grade-school to competition-level mathematical word problems. Solution depth varies from 2--10 steps depending on problem complexity. Verification is deterministic (numerical answer checking), but intermediate steps require symbolic manipulation and proper mathematical notation.

\item \textbf{Text-to-SQL}~\cite{scholak2021picard}: Generate SQL queries from natural language descriptions. Solutions must satisfy both syntactic constraints (valid SQL grammar) and semantic constraints (correct query semantics). Verification can be done via query execution on databases or schema validation.
\end{itemize}

\paragraph{D3. ~~Classical Planning.}
Structured domains with explicit action schemas and state transitions:

\begin{itemize}
\item \textbf{Blocksworld}~\cite{valmeekam2023planning,hao2023reasoning}: Rearrange blocks to achieve goal configurations using pick-and-place actions. Features deterministic state transitions, medium solution depth (5--15 steps), and external simulator-based verification. The challenge lies in managing preconditions and avoiding dead-ends.

\item \textbf{ALFWorld}~\cite{shridhar2021alfworld}: A text-based household environment with 134 evaluation tasks requiring goal achievement through sequences of high-level actions (e.g., \texttt{"go to place-X," "take object-Y from place-X"}). Agents receive binary rewards (1 for success, 0 otherwise). The main challenge is locating target objects and fulfilling household tasks using commonsense knowledge, often requiring 20--50 actions per task.

\item \textbf{BabyAI-Text}~\cite{carta2023babyai}: A grid-world environment extended from the BabyAI platform~\cite{chevalier2018babyai}, where agents and objects occupy an $8\times8$ tile room. Agents use 6 primitive actions (\texttt{turn left, turn right, go forward, pick up, drop, toggle}) to solve natural language tasks (e.g., \texttt{"Pick up the red box"}). Binary rewards (0 or 1) are provided. Tasks are challenging due to partial observability, obstacle avoidance, and the need for long-term planning described in natural language.
\end{itemize}

\paragraph{D4. ~~Interactive Web and E-Commerce.}
Domains involving dynamic environments with large action spaces:

\begin{itemize}
\item \textbf{WebShop}~\cite{yao2022webshop}: Agents purchase products based on natural language instructions (e.g., \texttt{"I need a long clip-in hair extension which is natural looking, and price lower than \$20.00"}) through web interactions such as search queries and button clicks. Unlike binary-reward environments, WebShop provides continuous rewards $[0, 1]$ based on how well the purchased product matches requirements. The action space is effectively unbounded due to free-form search queries and dynamic web results, requiring both search refinement and selection strategies.
\end{itemize}

\paragraph{D5. ~~Code Generation and Programming.}
Domains requiring syntactically valid and semantically correct program synthesis:

\begin{itemize}
\item \textbf{Programming}~\cite{zhou2023language,chen2021evaluating}: 
Generate code to solve programming tasks specified by natural language descriptions and test cases. Solutions can be arbitrarily deep (functions, classes, multi-file programs). Verification combines syntactic checking (parsing), semantic validation (type checking), and functional correctness (test case execution). The challenge lies in managing compositional structure and debugging failed attempts.
\end{itemize}

\paragraph{D6. ~~Open-Ended Creative Generation.}
Domains with subjective quality metrics and no unique correct solution:

\begin{itemize}
\item \textbf{Creative Writing}~\cite{yao2023tree}: Generate coherent stories or passages satisfying specified constraints (genre, themes, length). Solution quality is evaluated via LLM-based scoring or human judgment rather than deterministic verification. The challenge is balancing creativity, coherence, and constraint satisfaction without objective correctness criteria.
\end{itemize}

\paragraph{Domain Characteristics Summary.}
Table~\ref{tab:domain_characteristics} summarizes key characteristics of these benchmark domains, highlighting the diversity of challenges they present for ToT search methods.

\begin{table*}[t]
\centering
\begin{tabular}{@{}lcccc@{}}
\toprule
\textbf{Domain} & \textbf{Typical Depth} & \textbf{Branching} & \textbf{Verification} & \textbf{Primary Challenge} \\
\midrule
\multicolumn{5}{@{}c}{\textit{Domains with well-defined states and actions}} \\ \\
Game of 24 & 3 & High & Deterministic & Combinatorial search \\
Crosswords & 10 & Medium & Dictionary & Constraint satisfaction \\
Blocksworld & 5--15 & Low--Medium & Simulator & Precondition management \\
ALFWorld & 20--50 & Medium & Simulator & Object localization \\
BabyAI-Text & 10--30 & Low (6 actions) & Simulator & Partial observability \\
WebShop & 5--20 & Unbounded & Web Simulator & Dynamic action space \\
\midrule
\multicolumn{5}{@{}c}{\textit{Domains without well-defined states and actions}} \\ \\
Logical Inference & Variable & Medium & LLM & Deductive reasoning \\
Math Reasoning & 2--10 & Medium & Deterministic & Symbolic manipulation \\
Text-to-SQL & 30--100+ (Tokens) & High & Execution/Schema & Grammar + semantics \\
Creative Writing & 2--5 & Very High & LLM/Human & Subjective quality \\
Programming & Variable & Very High & Test execution & Compositional structure \\
\bottomrule
\end{tabular}
\caption{Characteristics of ToT benchmark domains. \textbf{Typical Depth} indicates solution length, \textbf{Branching} the number of viable successors per state, and \textbf{Verification} the primary method for validating solutions. Depth is a property of the problem graph and is well defined only for the domains in the upper block; for those in the lower block the entry describes the ToT as it is commonly constructed, and is a design choice rather than a property of the task. Branching is a count only where the action set is finite and enumerable, as with the six primitive actions of BabyAI-Text; where actions are free-form, as in WebShop, Creative Writing, and Programming, no count exists and the entry is qualitative. Depth is given in steps except for Text-to-SQL, whose search operates at token granularity (cf. Table~\ref{tab:design_choices}).}
\label{tab:domain_characteristics}
\end{table*}

These benchmarks collectively span a wide spectrum of search characteristics: from shallow, high-branching combinatorial problems (Game of 24) to deep, constrained sequential decision-making (ALFWorld, BabyAI-Text); from deterministic verification (puzzles, math) to subjective evaluation (creative writing); and from discrete action spaces (planning) to effectively unbounded spaces (WebShop, programming). This diversity makes ToT an ideal testbed for evaluating the generality and robustness of search algorithms across different problem structures.

\subsection{Common Evaluation Metrics}\label{sec:metrics} Across ToT studies, evaluation metrics generally fall into four categories:

\begin{enumerate}[label=\textbf{E\arabic*.},leftmargin=*] \item \textbf{Success Rate and Solution Quality (Effectiveness).} \label{eval:accuracy} This measures the ability of the search to find a valid solution. It is often subdivided into: \begin{itemize} \item \textit{Satisficing (Success Rate):} The binary percentage of problems where \textit{at least one} valid solution is found (e.g., Game of 24~\cite{yao2023tree}, Pass@1 in Coding~\cite{chen2021codex}). 

\item \textit{Optimality (Quality):} For tasks with graded outcomes (e.g., creative writing, optimizing code efficiency), this measures the scalar quality of the best solution found, often using human evaluation or model-based scoring~\cite{yao2023tree,hao2023reasoning}. \end{itemize}

\item \textbf{Search Efficiency (Computation Cost).} \label{eval:efficiency}
Quantifies the resources consumed to traverse the state space. Unlike classical search (which counts node expansions), LLM search costs are dominated by inference compute. This can be evaluated through:
\begin{itemize}
    \item \textit{Node Expansions:} The total number of thoughts generated or states visited~\cite{pendurkar2025policyguided}.
    \item \textit{Token Count:} The total number of LLM tokens generated~\cite{yao2023tree}.
\end{itemize}
Efficiency is often analyzed via \textbf{Pareto frontiers} (e.g., Accuracy vs. Tokens) to determine which search strategy yields the best return on computational investment.

\item \textbf{Solution Diversity and Robustness.} \label{eval:diversity}
Relevant for open-ended or ambiguous tasks where exploration is key. Metrics include:
\begin{itemize}
    \item \textit{Distinct Valid Paths:} The number of unique reasoning chains discovered that lead to a correct answer, a core metric in Self-Consistency~\cite{wang2023selfconsistency}.
    \item \textit{Entropy/Variance:} Measures of lexical or semantic diversity in the generated thoughts, ensuring the search does not collapse into a single mode~\cite{yao2023tree}.
\end{itemize}

\item \textbf{Heuristic Accuracy (Calibration).} \label{eval:calibration}
Unique to LLM-based search is the need to evaluate the \textit{search components} themselves. Since the heuristic function $h(s)$ is often a function approximator (e.g., a neural network), its reliability is critical.
\begin{itemize}
    \item \textit{Discriminative Accuracy:} How often the evaluator correctly identifies a valid vs. invalid partial state~\cite{shinn2024reflexion}.%,lightman2024lets}.
    \item \textit{Calibration Error:} The gap between the model's predicted confidence and its actual success rate. Poor calibration leads to ``delusional" search that pursues dead ends with high confidence.
\end{itemize}

\end{enumerate}

\section{Case Study: ToT in a Planning Domain}\label{sec:case_study}

To ground the formal definitions introduced in Section~\ref{sec:tot}, we present a concrete example illustrating how ToT is instantiated in a classical planning domain. The example demonstrates how a ToT node (representing a partial plan) is expanded using a specific proposal mechanism, evaluated using explicit $g$- and $h$-values, and validated via an external environment-based goal test.

\subsection{Domain and Problem Setup}

We consider a classical \emph{Blocksworld} planning domain~\cite{valmeekam2023planning} with given initial and goal configurations. We assume an external deterministic simulator capable of validating whether a proposed action sequence satisfies the goal condition, serving as the goal test $\mathcal{G}(s)$ (\ref{goal_test:external}).

\paragraph{State Representation.}
A state $s \in S$ is represented as a sequence of thoughts,
$s = [z_0, z_1, z_2, \dots, z_t]$, where $z_0$ encodes the problem description (the domain, initial, and goal configurations) and each thought $z_{i>0}$ describes a single action in natural language, e.g., \texttt{"pick block `A' from block `B' and place it on the table"}. 
The world state is implicitly tracked by the LLM's context window (and can also be explicitly tracked by an external simulator) by applying the action sequence to the initial configuration.

\paragraph{Root of ToT.}
The root node contains the state $s_0=[z_0]$, which consists only of the problem description. In our case study,

\begin{boxedquote}
$z_0 \gets $ \texttt{"In the Blocksworld domain [domain description]. The following plan will transform the initial configuration [initial config] into the goal configuration [goal config]."}
\end{boxedquote}

\paragraph{Cost Function ($g$).}
We consider the uniform cost function (\ref{cost:depth}), where each thought corresponds to a single action and incurs unit cost. Thus, $g(s) = t$ for state $s = [z_0, z_1, \dots, z_t]$.

\paragraph{Heuristic Function ($h$).}
The LLM is prompted to estimate the remaining distance to the goal, providing a scalar value (inverted \ref{h:value}). For example:

\begin{boxedquote}
\texttt{"Given the current plan" + [s] + ", provide an estimate of the minimal number of additional actions required to reach the goal configuration. Output a single scalar number."}
\end{boxedquote}

\paragraph{Evaluation Function ($f$).}
The evaluation function combines path cost and heuristic estimate:
\[
n.f = n.g + n.h.
\]
The $f$-value is used to prioritize ToT nodes during search.

\paragraph{Successor Function ($\mathcal{O}(s)$).}
Successors are generated using the following proposal mechanisms:
\begin{itemize}
\item \textbf{Independent sampling} (\ref{proposal_mechanism:sampling}) with top-$k=3$ sampling.
\item \textbf{Domain-specific constraints} (\ref{constraint:domain}), restricting outputs to valid Blocksworld operators (picking and placing blocks).
\end{itemize}

When expanding a ToT node with state $s$, the LLM is prompted:

\begin{boxedquote}
\texttt{[s] + "Generate the next valid planning step. Output a single Blocksworld action using the format: `pick block ? from ? and place it on ?'."}
\end{boxedquote}

\subsection{Example: Expanding a Node}

Consider a Blocksworld problem with three blocks $\{A, B, C\}$. The initial configuration is:
\begin{center}
\texttt{On(A,B)}, \texttt{On(B,C)}, \texttt{On(C,Table)}, \texttt{Clear(A)}
\end{center}
The goal configuration is:
\begin{center}
\texttt{On(C,B)}
\end{center}

Assume that during search, a node $n$ with the following state is reached:
\[
n.s = [z_0, z_1, z_2],
\]
where:
\begin{itemize}
\item $z_1 = $ \texttt{"Pick block `A' from `B' and place it on the table."} 
\item $z_2 = $ \texttt{"Pick block `B' from `C' and place it on block `A'."}
\end{itemize}

Note that $n.s$ implicitly defines a blocks configuration, but direct access to this configuration is not assumed. Moreover, the implied configuration may not be valid (e.g., attempting to pick a block that is not clear).

Before expanding $n$, the goal test $\mathcal{G}(n.s)$ is invoked by simulating the action sequence. In this example, the goal test returns \texttt{"False"}, so successor generation proceeds.

Using domain-constrained independent sampling with top-$k=3$, the LLM produces three candidate thoughts:
\begin{itemize}
\item $z_3^{(1)} = $ \texttt{"Pick block `A' from the table and place it on block `C'."} 
\item $z_3^{(2)} = $ \texttt{"Pick block `C' from the table and place it on block `B'."}
\item $z_3^{(3)} = $ \texttt{"Pick block `B' from block `A' and place it on the table."}
\end{itemize}

Each successor corresponds to a new state:
\[
s^{(i)} = [z_0, z_1, z_2, z_3^{(i)}].
\]

A state validator (Section~\ref{sec:goal_test}) can be used to check validity before evaluation. In this case, $s^{(1)}$ is invalid because block $A$ is not clear (block $B$ is on top of it). Consequently, $s^{(1)}$ is pruned and not added to the search tree.

\subsection{Evaluating Successors}
For each valid successor state, cost and heuristic estimates are computed. 

\paragraph{State $s^{(2)}$ (Move C onto B).}
The action sequence results in a configuration where $C$ is on $B$, satisfying the goal condition \texttt{"On(C,B)"}.
\[
g(s^{(2)}) = 3, \quad h(s^{(2)}) = 0 \text{ (Goal reached)}, \quad f(s^{(2)}) = 3
\]

\paragraph{State $s^{(3)}$ (Move B onto Table).}
This action moves $B$ away from $A$, but does not achieve \texttt{"On(C,B)"}. The LLM estimates at least one more step is needed to pick up $C$ and place it on $B$.
\[
g(s^{(3)}) = 3, \quad h(s^{(3)}) \approx 1, \quad f(s^{(3)}) = 4
\]

Under Best-First Search, the node with state $s^{(2)}$ has the minimal $n.f$ and is selected for expansion. The goal test $\mathcal{G}(s^{(2)})$ (via external simulator) returns \texttt{"True"}, and the sequence is returned as the solution.

\end{document}